\documentclass[10pt,twocolumn,letterpaper]{article}

\usepackage{cvpr}              
\usepackage{multirow}
\usepackage{url}
\usepackage{graphicx}
\usepackage{float}

\usepackage{tabularx}
\usepackage[table,xcdraw]{xcolor}
\usepackage[normalem]{ulem}
\useunder{\uline}{\ul}{}
\definecolor{cvprblue}{rgb}{0.21,0.49,0.74}
\usepackage[pagebackref,breaklinks,colorlinks,allcolors=cvprblue]{hyperref}

\def\paperID{*****} 
\def\confName{CVPR}
\def\confYear{2026}

\title{Diffusion MRI Transformer with a \\Diffusion Space Rotary Positional Embedding (D-RoPE)}

\author{
Gustavo Chau Loo Kung \quad
Mohammad Abbasi \quad
Camila Blank \quad
Juze Zhang \quad
Alan Q. Wang\\
Sophie Ostmeier\quad
Akshay Chaudhari \quad
Kilian Pohl \quad
Ehsan Adeli\\
\textit{Stanford University}
}

\begin{document}
\maketitle

\begin{abstract}

Diffusion Magnetic Resonance Imaging (dMRI) plays a critical role in studying microstructural changes in the brain. It is, therefore, widely used in clinical practice; yet progress in learning general-purpose representations from dMRI has been limited. A key challenge is that existing deep learning approaches are not well-suited to capture the unique properties of diffusion signals. 
Brain dMRI is normally composed of several brain volumes, each with different attenuation characteristics dependent on the direction and strength of the diffusion-sensitized gradients. Thus, there is a need to jointly model spatial, diffusion-weighting, and directional dependencies in dMRI. Furthermore, varying acquisition protocols (e.g., differing numbers of directions) further limit traditional models. 
To address these gaps, we introduce a diffusion space rotatory positional embedding (D-RoPE) plugged into our dMRI transformer to capture both the spatial structure and directional characteristics of diffusion data, enabling robust and transferable representations across diverse acquisition settings and an arbitrary number of diffusion directions. 
After self-supervised masked autoencoding pretraining, tests on several downstream tasks show that the learned representations and the pretrained model can provide competitive or superior performance compared to several baselines in these downstream tasks (even compared to a fully trained baseline); the finetuned features from our pretrained encoder resulted in a 6\% higher accuracy in classifying mild cognitive impairment and a 0.05 increase in the correlation coefficient when predicting cognitive scores. Code is available at: \href{github.com/gustavochau/D-RoPE}{github.com/gustavochau/D-RoPE}.
\end{abstract}
\vspace{-10pt}
\section{Introduction}

\begin{figure}[t]
    \centering
    \includegraphics[width=.9\linewidth]{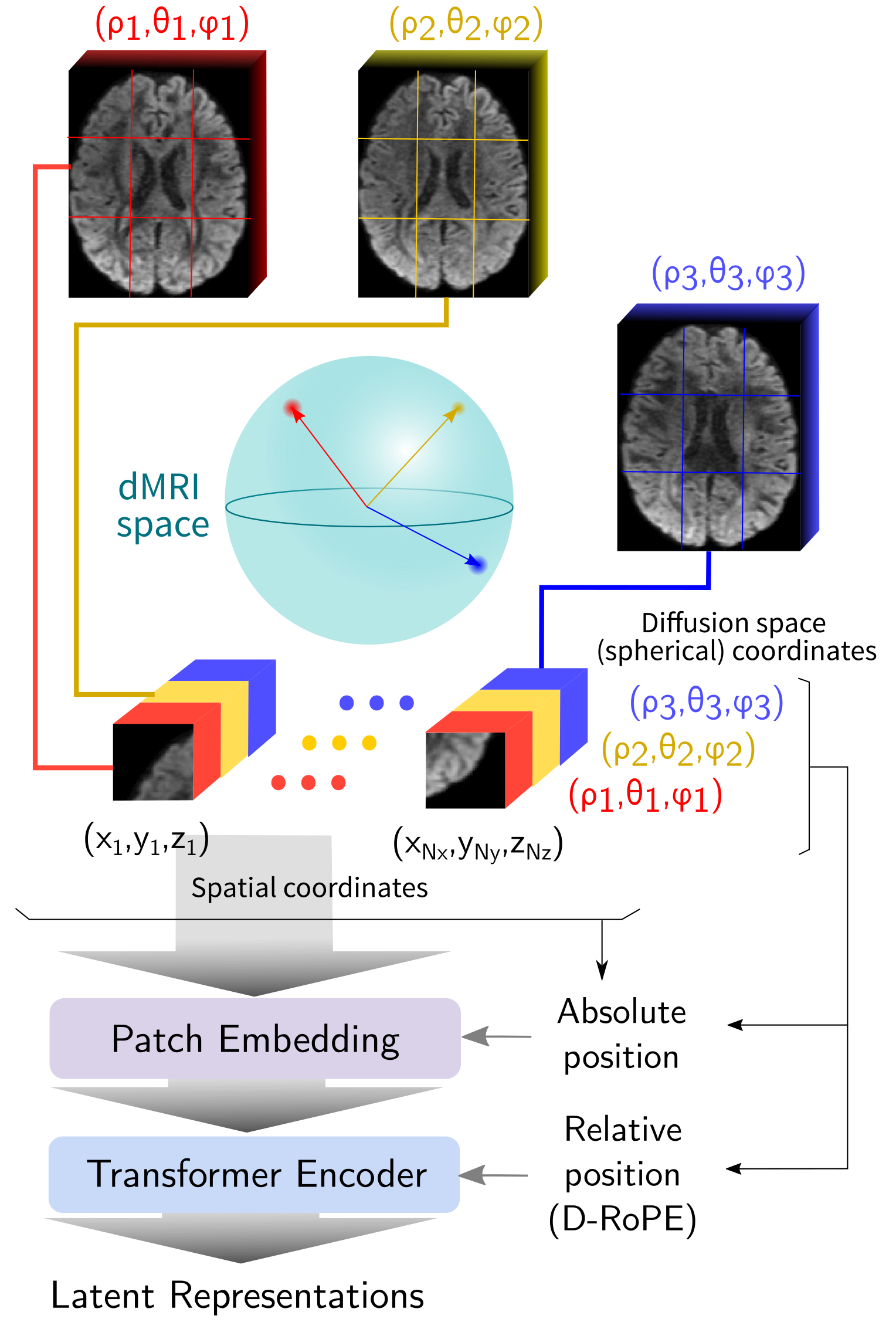} \vspace{-10pt}
    \caption{Overview of dMRI data structure and model input. The purple sphere represents diffusion sampling directions at a specific b-value. 3D dMRI volumes acquired at different directions are divided into 3D patches. Spatial coordinates $(x,y,z)$ define the image domain and spherical coordinates $(\rho,\theta,\varphi)$ describe the diffusion space.
    Patches are linearized with absolute positional embeddings and processed by a transformer encoder incorporating relative positional encoding (D-RoPE; see Section~\ref{sec:drope} and Fig.~\ref{fig:drope}).}
    \label{fig:intro}
    \vspace{-5pt}
\end{figure}

Diffusion Magnetic Resonance Imaging (dMRI) is a noninvasive imaging technique that is routinely used to probe the diffusion of water molecules inside the brain and other organs~\citep{basser1994mr}, and is widely used in clinical practice~\cite{drake2018clinical}. dMRI provides information about white matter integrity \citep{jones2013white, alexander2007diffusion}, putative structural connections \citep{mori1999three, johansen2009diffusion}, and tissue microstructure \citep{assaf2008diffusion, novikov2019quantifying}. In addition, metrics derived from this information are able to identify the effects of certain brain diseases, such as Alzheimer's \citep{pierpaoli2001diffusion, bozzali2002white} and multiple sclerosis \citep{filippi2001diffusion}. 
In dMRI, the molecular diffusion in a given voxel is encoded in the attenuation of the MRI signal and is dependent on the microstructure, the direction of the diffusion encoding gradient (normally called b-vector), and the ``strength'' of the diffusion encoding (normally expressed as a b-value). A normal pulsed gradient spin echo (PGSE) dMRI acquisition~\citep{Stejskal1965} can be viewed as a series of brain volumes, each corresponding to a point on the surface of a sphere. The diameter of the sphere comes from the b-value, while the position of the point in the sphere corresponds to the b-vector (see Figure \ref{fig:intro}).


While there have been extensive advances in novel architectures and representation learning for other medical imaging and MRI modalities~\citep{shinde2019predictive,liu2021handcrafted,qiang2020deep,caro2023brainlm},  
the progress in obtaining general-purpose representations from dMRI has been limited~\citep{huang2025autoencoder, singh2025interpretable}, or the representations have only focused on specific applications~\citep{wang2025self, xiao2024samrobnoddi}. A key challenge is that existing deep learning approaches are not well-suited to capture the unique properties of the spatial information and directional dependencies in diffusion signals. Additionally, the problem is further complicated by the multiple and varying diffusion protocols (e.g., different numbers of diffusion encoding directions and b-values) often used to acquire dMRI.
Previous attempts at dealing with these challenges for specific applications include: upsampling to a common protocol~\cite{gibbons2019simultaneous}, modeling the diffusion space as a graph~\cite{chen2022hybrid, yang2023towards, chen2020estimating}, using modified convolutional operations~\cite{lyon2023spatio}, or concatenating the protocol information as part of the input~\citep{zong2024attention}.


As a result, there is a need for architectures that can model the joint spatial and diffusion space dependencies of dMRI and that can handle a variety of dMRI protocols to obtain general and transferable representations. In this paper, we propose a model that combines novel absolute and relative positional embeddings to encode the geometric information of the different volumes (Figure \ref{fig:intro}). 
The proposed model is able to handle acquisitions with multiple b-values and an arbitrary number of directions. 

In summary, the contributions of this work are as follows: (i) we propose a novel architecture inspired by video transformers that can obtain representations from arbitrary dMRI acquisitions; (ii) we propose D-RoPE, which augments the attention mechanism with a generalization of Rotary Positional Embedding (RoPE) that models the interdependence of different diffusion directions and b-values; and (iii) we perform a thorough evaluation of the architecture and test the representations in several downstream tasks (Age regression, sex classification, mild cognitive impairment (MCI) classification, and Alzheimer’s disease assessment scale (ADAS) regression). 

\section{Related Work}

\textbf{Task-specific models for diffusion MRI.}
Some previous work focused on obtaining high-resolution reconstructions of the diffusion tensor ~\citep{karimi2022diffusion}, or of maps of diffusion tensor metrics (e.g. fractional anisotropy, mean diffusivity) or other microstructural parameters ~\citep{zheng2023microstructure,zong2024attention}. 
Other works have focused on obtaining meaningful information from the dMRI images or dMRI-derived connectomes ~\citep{chen2022hybrid,yang2023towards,chen2024deep}.  
Though these and other studies have shown the effectiveness of deep learning for dMRI, there is a need to develop models that can be adapted for a wide variety of tasks. 

\noindent\textbf{Representation Learning for Brain MRIs.}
Foundation models have been proposed to extract representations of structural (T1 and T2 weighted) brain MRI~\citep{tak2024foundation, barbano2024anatomical}. Another recent publication combined structural MRI and diffusion MRI for different tasks ~\citep{karimi2024approach}, including tractography streamline generation. Foundation models have also been proposed for other modalities, like functional MRI (fMRI)~\citep{caro2023brainlm,dong2024brain}. For dMRI, existing representation learning work has focused on superresolution~\citep{wang2025self}, microstructure reconstruction~\citep{xiao2024samrobnoddi}, and tractography features or representations~\citep{singh2025interpretable,huang2025autoencoder}, but there has been a lack of work focusing on obtaining general and transferable representations from raw dMRI data.

\noindent\textbf{Attention in Video Transformers}
There have been different strategies for adapting attention modules for video inputs, so as to simultaneously handle attention in the space and time domains. \citep{bertasius2021space} showed one of the first attempts by alternating attention in space and time. Prior works looked at alternative ways of integrating both attentions with reduced complexity via space-time mixing~\citep{bulat2021space}, or via tubelets and factorized attention~\citep{arnab2021vivit}. 

\noindent\textbf{Rotary Positional Embedding}
\cite{su2024roformer} introduced the concept of Rotary Positional Embedding (RoPE) to integrate information about the relative position of tokens in the computation of the attention scores. Since then, some works have looked at different ways of adapting RoPE to the multidimensional case of video~\citep{wang2024qwen2,wei2025videorope}. Recently, LieRE~\citep{ostmeier2024liere} was proposed to generalize RoPE to high-dimensional rotation matrices for 2D and 3D vision tasks. Despite all this progress, none of this relative positional information can be applied directly to dMRI data, which is normally represented in a spherical space. 

\section{Methods}

\begin{figure}[t]
    \centering
    \includegraphics[width=1\linewidth]{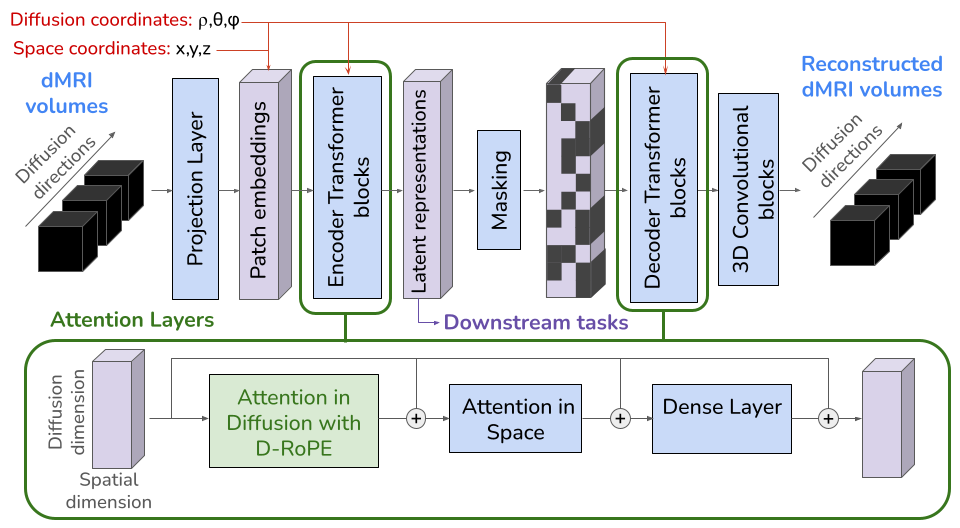}
    \caption{Summary of the model architecture. The different diffusion volumes are patchified via a projection layer and the absolute positional information of the image space and the diffusion space is added to the the patch embeddings. 
    The tokens are passed through modified attention blocks that alternate attention in both image and diffusion spaces. Additionally, relative information of different diffusion directions and b-values is encoded via D-RoPE. The obtained latent representations from the encoder are used for downstream task evaluation. To close the pretraining loop, some tokens are masked and fed into transformer blocks as the ones used in the encoder, followed by 3D convolutional blocks to reconstruct the original volumes.}
    \label{fig:archi}
    \vspace{-10pt}
\end{figure}

We adopt a transformer model architecture and train it with the Masked Autoencoding (MAE)~\citep{he2021mae} objective as summarized in Figure~\ref{fig:archi}. Given a series of $N_d$ diffusion volumes of size $(N_x,N_y,N_z)$, the input is divided into 3D patches of size $(P_x,P_y,P_z)$, as shown in Figure~\ref{fig:intro}, and then linearized to tokens of latent dimension $d$. Thus, the resulting patch embedding is of size $S \times N_d \times d$, where $S = \frac{N_x}{P_x} \times \frac{N_y}{P_y} \times \frac{N_z}{P_z}$.

Vision Transformer (ViT)~\citep{dosovitskiy2020image} architectures typically use a learnable positional encoding~\cite{karimi2022diffusion,zheng2023microstructure} or a fixed sinusoidal positional encoding~\citep{dosovitskiy2020image}; however, this approach will be inadequate for samples with different dMRI protocol information. In our model, the absolute positional encoding is modified to include both spatial and diffusion-space information. The spatial position $(x,y,z)$ is encoded using a 3D sinusoidal positional encoding into a $d/2$ dimensional vector $p_s$. The diffusion direction information, expressed in spherical coordinates $(\rho,\theta,\varphi)$, is encoded using a learnable linear layer into another $d/2$ dimensional vector $p_d$. Both vectors, $p_s$ and $p_d$, are concatenated and then summed to the patch embeddings. 
The embeddings are fed to the transformer encoder, which consists of a modified ViT~\citep{dosovitskiy2020image}. Inspired by video transformers~\citep{bertasius2021space}, the attention blocks alternate between the spatial and diffusion dimensions. 
The attention blocks also include a modified relative positional encoding (details in Section~\ref{sec:drope}), which enables the model to exploit the interdependence between the different diffusion volumes. 
With this structure of the positional embedding and the attention blocks, the encoder supports an arbitrary number of diffusion volumes. 

The encoder output provides the latent representation used for downstream tasks. To close the pretraining loop, some tokens are masked and fed into transformer blocks as the ones used in the encoder, followed by 3D convolutional blocks to reconstruct the original volumes. The loss function is a combination of the Mean Squared Error (MSE) of the masked and unmasked patches:
\begin{equation}
    L = (1-\tau) L_\text{unmasked} + \tau L_\text{masked},
\end{equation}
where $\tau$ is linearly ramped from 0.05 to 0.95 across epochs.

\subsection{D-RoPE}
\label{sec:drope}
The original Rotary Positional Embedding (RoPE)~\cite{su2024roformer} modifies the attention mechanism by applying a rotation matrix $R_{\alpha, m-n}^d$ to encode the exact relative distance $m-n$ between the $m$ th query, $q_m$, and the $n$-th key, $k_n$ :
\begin{equation}
\resizebox{0.9\columnwidth}{!}{$
\mathbf{R}^{d}_{\alpha, \Delta} =
\left(
\begin{array}{cccccc}
\cos \Delta\alpha_1 & -\sin \Delta\alpha_1 &  \cdots & 0 & 0 \\
\sin \Delta\alpha_1 & \cos \Delta\alpha_1  & \cdots & 0 & 0 \\
0 & 0 &  \ddots & \vdots & \vdots \\
\vdots & \vdots &  \ddots & \cos \Delta\alpha_{d/2} & -\sin \Delta\alpha_{d/2} \\
0 & 0 & \cdots & \sin \Delta\alpha_{d/2} & \cos \Delta\alpha_{d/2}
\end{array}
\right)$,}
\label{eq:rope}
\end{equation}
where  $\Delta=m-n$, $d$ is the number of embedding dimensions, and $\alpha_i = 10000^{-2(i-1)/d}, i \in [1, 2, \ldots , d/2]$. The similarity between $q_m$ and $k_n$ involves the inner product: 
\begin{equation}
\langle q_m, k_n \rangle= {q_m}^\top \mathbf{R}^{d}_{\alpha, m-n}  k_n.
\end{equation}
Due to the block diagonal structure of  $\mathbf{R}^{d}_{\alpha, m-n}$ with 2x2 rotations blocks, efficient implementations apply element-wise rotations directly to the query vector $x_m$ and key vectors $x_n$. 

We can think of $\mathbf{R}^{d}_{\alpha, m-n}$ as encoding relative distances, $(m-n)$ of a 1D sequence. However, dMRI volumes have a more complex spatial structure: each 3D volume is associated with a point $(b,\mathbf{v}) \in \mathbb{R}^+ \times \mathbb{S}^2$ in the diffusion space, where $b$ is the b-value and $\mathbf{v}$ is the diffusion gradient direction (b-vector), and such the linear and ordinal distance used in RoPE is not appropriate for this case. 
To generalize RoPE, we substitute $(m-n)$ by a specifically designed distance $D(m,n)$. 
As a result, (\ref{eq:rope}) becomes (the $m$ and $n$ dependency is omitted for clarity):
\begin{equation}
\resizebox{0.9\columnwidth}{!}{$
\mathbf{R}^{d}_{\theta, m, n} =
\left(
\begin{array}{cccccc}
\cos D\alpha_1 & -\sin D\alpha_1 &  \cdots & 0 & 0 \\
\sin D\alpha_1 & \cos D\alpha_1  &  \cdots & 0 & 0 \\
0 & 0 &  \ddots & \vdots & \vdots \\
\vdots &  \ddots & \ddots & \cos D\alpha_{d/2} & -\sin D\alpha_{d/2} \\
0 & 0 &  \cdots & \sin D\alpha_{d/2} & \cos D\alpha_{d/2}
\end{array}
\right).$}
\label{eq:drope}
\end{equation}

\begin{figure*}[h]
    \centering
    \includegraphics[width=\linewidth]{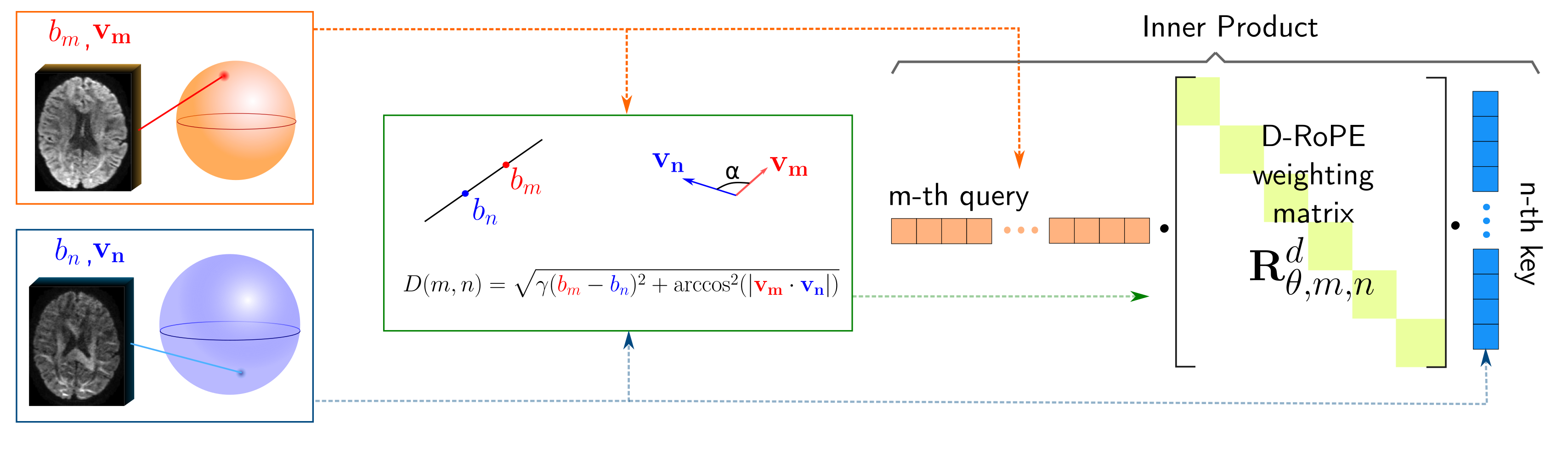}
    \caption{Attention calculation with D-RoPE: The relative rotation matrix depends on the distance between diffusion volumes, which is defined as a weighted combination of the distance between b-values and the angle between b-vectors.}
    \label{fig:drope}
    \vspace{-10pt}
\end{figure*}

The distance $D(m,n)$ models the geometry of the diffusion space. We consider that two tokens $x_m$ and $x_n$ correspond to two diffusion acquisitions represented by $(b_n,\mathbf{v_n}), (b_m,\mathbf{v_m}) \in \mathbb{R}^+ \times \mathbb{S}^2$. That is, the corresponding b-values are defined in a linear space and the b-vectors are defined on the unitary sphere surface. Then, we define $D(m,n)$ as:
\begin{align}
         D(m,n)&=D\big((b_m, \mathbf{v}_m), (b_n, \mathbf{v}_n)\big) \\
         &=\sqrt{\gamma(b_m-b_n)^2 + \arccos^{2}(|\mathbf{v_m} \cdot \mathbf{v_n}|)},
\end{align}

\noindent where $\gamma$ is a non-negative parameter that allows for weighting the contributions of the different components. Note that the absolute value inside the arc cosine provides 180° symmetry, where opposite gradient directions $(\mathbf{v}$ and $-\mathbf{v}$) are treated as identical, since they produce equivalent diffusion encoding. The concept of this diffusion space rotary positional embedding (D-RoPE), is summarized in Figure~\ref{fig:drope}.

\subsection{Datasets and preprocessing}

To train and evaluate our model, we used the combination of datasets from the Human Connectome Project (HCP) \citep{elam2021human,somerville2018lifespan} and the human Alzheimer’s Disease Neuroimaging Initiative (ADNI)~\citep{mueller2005alzheimer}. 
First, the HCP Young Adult (HCP-YA)~\citep{elam2021human} dataset (n=1,065) was used for the pretraining stage and evaluation of the MAE reconstruction task with a 70/15/15 split for training/validation/test. 
For downstream evaluation, HCP Development (HCP-D, n=612)~\citep{somerville2018lifespan}, HCP Aging (HCP-A, n=708)~\citep{bookheimer2019lifespan}, and the fourth phase of ADNI (n=276)~\citep{mueller2005alzheimer} were used under a cross-validation setting. Experiments are detailed in Section~\ref{sec:downstream}.

For HCP-A, HCP-D, and ADNI, distortion correction was conducted using Topup~\citep{andersson2003topup,smith2004advances} followed by Eddy current correction~\citep{andersson2016eddy} using our publicly available preprocessing pipeline~\citep{abbasi2025dti} and FSL~\citep{jenkinson2012fsl}.
HCP-YA data already include these two steps~\cite{glasser2013minimal}. In case of multiple scans per subject, the different scans were rigidly registered and merged. Finally, the merged dMRI volumes were registered by a combination of rigid and affine transformations to MNI space using ANTs~\citep{avants2008ants} to account for possible confounders of brain size and volume. To be input to the network, the diffusion weighted images were normalized by the b=0 reference volume, and outliers due to numerical errors were clipped (No logarithmic transformation was applied).

\section{Experiments}

\subsection{Masked AutoEncoder Pretraining}

We considered three different masking strategies: (i) normal spatial masking ratio of 75\%, (ii) masking in the diffusion direction (i.e., random directions are entirely masked) with a ratio of 50\%, and (iii) an alternation of the previous two on an epoch-by-epoch basis. In the following subsections, they are referred to as spatial, diffusion, and alternating, respectively. 
Ten encoding layers and three decoding layers were used, followed by three 3D convolutional layers, with an embedding dimension size of 384. The MAE was trained with a learning rate and weight decay following a cosine schedule. 
A patch size of (8,8,4) was used for all experiments (An evaluation of other patch sizes is shown in section~\ref{sec:add_eval} in the supplementary material). At each epoch, slabs of four contiguous slices and 15 random directions are selected for each patient to reduce memory consumption and act as a form of data augmentation. A detailed table with the hyperparameters used during the pretraining stage is shown in Table \ref{tab:pret_param} in the supplementary material.

\subsection{Ablation and Evaluation}

In addition to the three different versions of masking strategies (spatial, diffusion, and alternating), we evaluated both the MAE with and without D-RoPE. We compared reconstructed volumes against the ground-truth volumes in the test set after a 75\% spatial masking in 90 randomly selected directions. The reconstructions were compared in terms of Peak signal-to-noise-ratio (PSNR), Structural Similarity Image Metric (SSIM), and Fréchet Inception distance (FID)~\cite{heusel2017gans}. For the FID, the features of the different layers of a pretrained 2D ResNet50~\cite{he2015deep} architecture were used.   
Additionally, we fitted the diffusion tensor model to the reconstructed volumes of each method and obtained fractional anisotropy (FA) and mean diffusivity (MD) maps using DIPY~\cite{garyfallidis2014dipy}. We compared the mean absolute error between these fitted maps and the ground-truth maps in several known white matter tracts (e.g., corpus callosum, inferior longitudinal fasciculus, etc.) to test the anatomical plausibility of the generated images. 
The segmentation of the tracts was obtained using TractSeg~\citep{wasserthal2018tractseg} on the ground-truth data.

\subsection{Downstream Evaluations}
\label{sec:downstream}

We tested the learned representations on the following downstream tasks: 
\begin{enumerate}
    \item \textbf{Brain Age Prediction:} We used a combination of HCP-D and HCP-A. The age was z-score normalized. 
    \item \textbf{Sex classification:} We decided to use only HCP-A as the sex differences might be more obvious in the early developmental stage. 
    \item \textbf{Cognitive normal (CN) versus mild cognitive impairment (MCI) classification:} This was performed in the ADNI dataset. As 30\% of the samples are MCI, we used a positive sample weighting in the cross-entropy loss. 
    \item \textbf{Prediction of Alzheimer’s disease assessment scale--cognitive subscale (ADAS-Cog):} The ADAS-Cog is a brief neuropsychological assessment to evaluate the severity of cognitive symptoms associated with dementia~\citep{kueper2018alzheimer}. This task was performed in the ADNI dataset. The score was z-score normalized. 
\end{enumerate}

For all downstream tasks, we performed five-fold cross-validation. HCP-D and HCP-A~\citep{bookheimer2019lifespan} were divided into train/val/test splits with a 30/35/35 ratio. ADNI was divided using a 60/20/20 ratio, and the splits were stratified to maintain the same proportion of CN vs MCI subjects. 
For all downstream tasks, we used the pretrained model with D-RoPE and alternating masking, which achieved the best reconstruction performance. The hyperparameters used are detailed in Table~\ref{tab:ft_param} in the supplementary material.
 
We tested three approaches 1) fine-tuning the last encoder layer of the model without the decoder and adding a prediction head 2) linear probing on the frozen representations 3) an MLP head on top of the frozen representations. 

All of these variants have as input the CLS token from the encoder (384-dimensional vector), while the input to the encoder are dMRI volumes cropped to a size of (192,224,56).
As done previously~\cite{zong2024attention}, to show that our model can handle varying diffusion directions and b-values, we randomly sample 12 diffusion directions/b-value pairs per subject. This effectively corresponds to having a separate acquisition protocol per subject. A simpler comparison with just two different fixed protocols is included in Section~\ref{sec:add_eval} in the supplementary materials. 

We compared our pretrained MAE with several baselines. We included a 3D ResNet18 (Which was used previously to regress age from DTI maps~\cite{gao2024predicting}) trained on the FA and MD maps in a fully supervised manner. The FA and MD maps were computed using DIPY~\cite{garyfallidis2014dipy} from the same 12 directions that were selected as input to our model. Additionally, we considered a standard ViT encoder (Adapted from~\cite{zheng2023microstructure}) with one of two inputs: the aforementioned FA and MD maps or the 12 randomly selected dMRI directions. For the latter, the different diffusion directions are considered as channels in the projection layer. The ViT included a 3D version of regular RoPE~\citep{su2024roformer}, which operates on the image space. The number of encoder layers (10) and embedding size (384) was chosen to match our pretrained MAE. 
Additionally, we included a comparison with typically used handcrafted features in dMRI~\cite{wen2021reproducible}. We extracted the mean FA and MD values for different notable white matter tracts (Tracts were obtained using TractSeg~\citep{wasserthal2018tractseg}) and concatenated the values into a 130-dimensional vector. Predictions were then obtained by either doing linear probing on top of this vector or adding an MLP head on top of it. 
Additionally, a multi-tissue multi-shell model~\cite{jeurissen2014multi} was used to estimate spherical harmonic (SH) coefficients of white matter up to the 4th order using MRtrix3~\cite{tournier2019mrtrix3}. Rotation-invariant spherical harmonic (RISH) features were then computed by summing the squared magnitudes (power) of the SH coefficients within each harmonic order. We computed the mean value for each harmonic order along the Tractseg segmented white matter tracts and concatenated the values into a 165-dimensional vector. Predictions were then obtained by adding an MLP head on top.

\section{Results}

\subsection{MAE Reconstruction}

\begin{figure*}[h]
    \centering
    \includegraphics[width=0.95\linewidth]{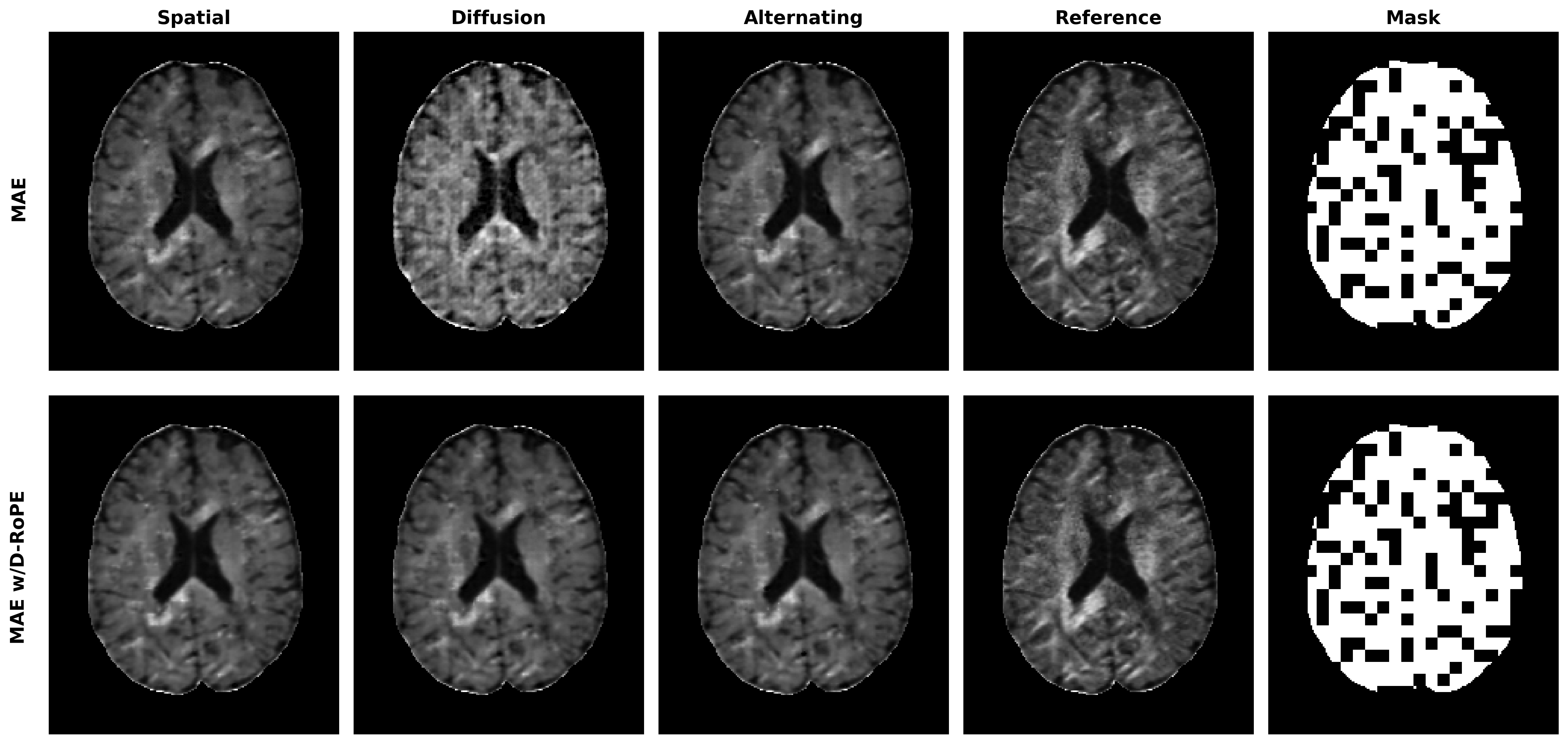}
    \caption{Qualitative evaluation of the reconstructions for a test subject for a b-value of 2000 $s^2/mm$. The reconstruction obtained with different masking strategies (Spatial, Diffusion, and Alternating) and with and without D-RoPE are shown in the three left-most columns. The reference ground-truth image and the spatial mask that was used are shown in the two right-most columns (white=masked). In general, a good reconstruction of the brain structures are observed. The MAE without D-RoPE case shows an erroneous contrast when only diffusion space masking is performed, possibly because of the lack of spatial information for a given diffusion direction.}
    \label{fig:b2000_recon}
    \vspace{-5pt}
\end{figure*}

Exemplary reconstructions from the different masking strategies (Spatial, Diffusion, and Alternating) are shown in Figure~\ref{fig:b2000_recon} for one of the volumes with a b-value of 2000 $s^2/mm$. We compare the MAE with and without D-RoPE. Additional reconstructions for b-values of 1000 and 3000 $s^2/mm$ are shown in the supplementary material (Figures~\ref{fig:b1000_recon} and~\ref{fig:b3000_recon}). We observe a good reconstruction performance for most methods; however, the version without D-Rope and diffusion space masking captures the structure well but not the general intensities. This is possibly due to the limited information it learns about the relationships between the different directions and b-values.

The quantitative evaluation of the different reconstructions is shown in terms of PSNR, SSIM, and FID in Table~\ref{tab:rec_eval_psnr}. The evaluation using the DTI metrics derived from the reconstructions is shown in Table~\ref{tab:rec_eval_fa}. We observe that D-RoPE with diffusion masking and D-RoPE with alternating masking tend to provide better metrics overall (e.g. reductions of 37-40\% in FID when comparing the later against MAE with Spatial masking). We also note that alternating masking helps the MAE without D-RoPE case to improve performance, possibly because it forces the model to combine more information about the spatial and diffusion spaces. Additionally, the low error in MD and relatively low error in FA suggest that the reconstructions are able to maintain the general microstructural features contained in the original dMRI volumes. 

\begin{table*}[h]
\caption{Quantitative evaluation of the reconstructions obtained from different masking strategies (Spatial, Diffusion, and Alternating) and with and without D-RoPE in the test set of HCP-YA. The mean $\pm$ standard deviation is shown for Peak Signal-to-Noise Ratio (PSNR) and Structural Similarity Index Measure (SSIM). The value of the Fréchet Inception distance (FID) is shown in the last column. $\uparrow$: the higher, the better; $\downarrow$: the lower, the better. The best results per b-value are highlighted in bold.}
\label{tab:rec_eval_psnr}
\centering
\renewcommand{\arraystretch}{1.25} 
\begin{tabular}{c|ccccc}
\hline
 & Method & Masking & PSNR ($\uparrow$) & SSIM ($\uparrow$) & FID ($\downarrow$) \\ \hline
\multirow{6}{*}{b-value = 1000 $s^2/mm$} & MAE & Spatial & 14.54 ± 0.16 & 0.971 ± 0.003 & 13.37 \\
 & MAE & Diffusion & 10.13 ± 0.26 & 0.875 ± 0.005 & 39.76 \\
 & MAE & Alternating & 14.70 ± 0.17 & 0.971 ± 0.004 & 11.13 \\ \cline{2-6} 
 & MAE w/D-RoPE & Spatial & 14.84 ± 0.17 & 0.973 ± 0.003 & 10.28 \\
 & MAE w/D-RoPE & Diffusion & 14.86 ± 0.17 & \textbf{0.974 ± 0.003} & 10.73 \\
 & MAE w/D-RoPE & Alternating & \textbf{14.94 ± 0.17} & \textbf{0.974 ± 0.003} & \textbf{8.36} \\ \hline
\multirow{6}{*}{b-value = 2000 $s^2/mm$} & MAE & Spatial & 14.91 ± 0.17 & 0.969 ± 0.003 & 13.49 \\
 & MAE & Diffusion & 10.07 ± 0.28 & 0.844 ± 0.006 & 39.86 \\
 & MAE & Alternating & 15.07 ± 0.18 & 0.968 ± 0.004 & 10.81 \\ \cline{2-6} 
 & MAE w/D-RoPE & Spatial & 15.26 ± 0.17 & 0.972 ± 0.002 & 9.61 \\
 & MAE w/D-RoPE & Diffusion & 15.31 ± 0.18 & \textbf{0.973 ± 0.002} & 9.99 \\
 & MAE w/D-RoPE & Alternating & \textbf{15.37 ± 0.18} & \textbf{0.973 ± 0.002} & \textbf{8.05} \\ \hline
\multirow{6}{*}{b-value = 3000 $s^2/mm$} & MAE & Spatial & 15.13 ± 0.17 & 0.969 ± 0.003 & 14.97 \\
 & MAE & Diffusion & 10.01 ± 0.30 & 0.819 ± 0.009 & 41.57 \\
 & MAE & Alternating & 15.29 ± 0.18 & 0.968 ± 0.003 & 11.54 \\ \cline{2-6} 
 & MAE w/D-RoPE & Spatial & 15.49 ± 0.17 & 0.971 ± 0.002 & 10.42 \\
 & MAE w/D-RoPE & Diffusion & 15.54 ± 0.18 & \textbf{0.973 ± 0.002} & 10.63 \\
 & MAE w/D-RoPE & Alternating & \textbf{15.61 ± 0.18} & \textbf{0.973 ± 0.002} & \textbf{8.95} \\ \hline
\end{tabular}
\end{table*}

\begin{table}[h!]
\caption{Evaluation of fractional anisotropy (FA) and mean diffusivity (MD) obtained from the reconstructions with and without D-RoPE in HCP-YA. For MD, the x1000 error is shown. The mean $\pm$ standard deviation of the absolute error is shown. $\downarrow$: the lower, the better. The best results are highlighted in bold.}
\label{tab:rec_eval_fa}
\renewcommand{\arraystretch}{1.15} 
\resizebox{\linewidth}{!}{%
\centering
\begin{tabular}{cccc}
\hline
Method & Masking & FA error ($\downarrow$) & MD error x$10^3$($\downarrow$) \\ \hline
MAE& Spatial & 0.074 ± 0.013 & 0.098 ± 0.009 \\
MAE & Diffusion & 0.185 ± 0.012 & 0.233 ± 0.015 \\
MAE & Alternating & 0.080 ± 0.013 & 0.109 ± 0.008 \\ \hline
MAE w/D-RoPE & Spatial & 0.061 ± 0.007 & 0.104 ± 0.008 \\
MAE w/D-RoPE & Diffusion & 0.058 ± 0.007 & \textbf{0.099 ± 0.008} \\
MAE w/D-RoPE & Alternating & \textbf{0.055 ± 0.007} & 0.103 ± 0.008 \\ \hline
\end{tabular}
}
\vspace{-10pt}
\end{table}

\subsection{Downstream Evaluation}

\begin{table*}[h!]
\caption{Evaluation of different methods (handcrafted features, 3D ResNet, ViT, and our partially supervised methods) on the age prediction and sex classification tasks in HCP-A and HCP-D. For age prediction, the correlation coefficient ($\rho$) and mean square error (MSE) are reported. For sex classification, accuracy (ACC) and area under the receiver operating curve (AUROC) are reported. $\uparrow$: the higher, the better, $\downarrow$: the lower, the better. The best result per metric is highlighted in bold, and the second best is underlined. Our proposed methods are highlighted in yellow. *HC Features means Hand-Crafted Features.
}
\label{tab:hcpagdev_metrics}
\vspace{-10pt}
\renewcommand{\arraystretch}{1.2} 
\resizebox{\textwidth}{!}{%
\centering
\begin{tabular}{ccccccccc}
 &  &  &  & \multicolumn{2}{c}{Age prediction} &  & \multicolumn{2}{c}{Sex classification} \\ \cline{5-9} 
Method & Features & \begin{tabular}[c]{@{}c@{}}Prediction \\ Head\end{tabular} & Supervision & $\rho$ ($\uparrow$) & MSE ($\downarrow$) & \textbf{} & ACC ($\uparrow$) & AUROC ($\uparrow$) \\ \hline
HC Features & Tract FA, MD & Linear & Full & 0.90 ± 0.01 & 0.20 ± 0.01 &  & 70.90 ± 2.60 \% & 0.77 ± 0.02 \\
HC Features & Tract FA, MD & MLP & Full & 0.90 ± 0.01 & 0.19 ± 0.02 &  & 56.90 ± 2.50 \% & 0.59 ± 0.03 \\
HC Features & Tract based RISH & MLP & Full & 0.61 ± 0.21 & 1.15 ± 1.24 & \multicolumn{1}{l}{} & 61.70 ± 2.00 \% & 0.67 ± 0.03 \\ \hline
3D ResNet & FA and MD maps & MLP & Full & \textbf{0.96 ± 0.01} & \textbf{0.09 ± 0.02} & {\ul } & \textbf{76.00 ± 1.20 \%} & \textbf{0.83 ± 0.04} \\
\multicolumn{1}{l}{ViT w/RoPE} & FA and MD maps & MLP & Full & 0.93 ± 0.01 & {\ul 0.15 ± 0.02} &  & 65.40 ± 3.70 \% & 0.72 ± 0.04 \\
\multicolumn{1}{l}{ViT w/RoPE} & Raw dMRI & MLP & Full & 0.89 ± 0.02 & 0.34 ± 0.11 &  & 65.10 ± 2.60 \% & 0.71 ± 0.02 \\ \hline
\rowcolor[HTML]{FAFADC} 
Ours & Frozen latent & Linear & Head & 0.89 ± 0.01 & 0.24 ± 0.02 & \textbf{} & {\ul 74.60 ± 2.90 \%} & {\ul 0.82 ± 0.02} \\
\rowcolor[HTML]{FAFADC} 
Ours & Frozen latent & MLP & Head & 0.90 ± 0.02 & 0.20 ± 0.03 &  & 59.30 ± 2.30 \% & 0.64 ± 0.02 \\
\rowcolor[HTML]{FAFADC} 
Ours & Last layer finetuned & MLP & Partial & {\ul 0.94 ± 0.01} & {\ul 0.15 ± 0.03} &  & 69.60 ± 3.10 \% & 0.77 ± 0.02
\end{tabular}
}
\end{table*}

\begin{table*}[h!]
\caption{Evaluation of different methods (handcrafted features, 3D ResNet, ViT, and our partially supervised methods) on the CN vs MCI classification and ADAS prediction tasks in ADNI. For CN vs MCI, balance accuracy (BA), area under the precision-recall curve (AUPRC), and area under the receiver operating curve (AUROC) are reported. For ADAS prediction, the correlation coefficient ($\rho$) and mean square error (MSE) are reported. $\downarrow$: the lower, the better. $\downarrow$: the lower, the better. The best result per metric is highlighted in bold, and the second best is underlined. Our proposed methods are highlighted in yellow. *HC Features means Hand-Crafted Features.}
\label{tab:adni_metrics}
\vspace{-8pt}
\renewcommand{\arraystretch}{1.2} 
\resizebox{\textwidth}{!}{%
\begin{tabular}{cccccccccc}
 &  &  &  & \multicolumn{3}{c}{CN vs MCI classification} &  & \multicolumn{2}{c}{ADAS prediction} \\ \cline{5-10} 
Method & Features & \begin{tabular}[c]{@{}c@{}}Prediction\\ Head\end{tabular} & Supervision & BA ($\uparrow$) & AUPRC ($\uparrow$) & AUROC ($\uparrow$) &  & $\rho$ ($\uparrow$) & MSE ($\downarrow$) \\ \hline
HC Features & Tract FA, MD & Linear & Full & {\ul 61.90 ± 7.30 \%} & \textbf{0.53 ± 0.06} & \textbf{0.69 ± 0.06} &  & 0.06 ± 0.15 & 10.92 ± 4.08 \\
HC Features & Tract FA, MD & MLP & Full & 57.40 ± 5.40 \% & 0.44 ± 0.09 & 0.62 ± 0.04 &  & {\ul 0.33 ± 0.09} & {\ul 0.81 ± 0.13} \\
HC Features & Tract based RISH & MLP & Full & 53.90 ± 9.60 \% & 0.31 ± 0.07 & 0.52 ± 0.08 & \multicolumn{1}{l}{} & 0.28 ± 0.14 & 0.93 ± 0.18 \\ \hline
3D ResNet & FA and MD maps & MLP & Full & 58.40 ± 8.30 \% & {\ul 0.50 ± 0.10} & {\ul 0.68 ± 0.08} &  & {\ul 0.33 ± 0.08} & 0.86 ± 0.14 \\
ViT w/RoPE & FA and MD maps & MLP & Full & 58.10 ± 6.60 \% & 0.47 ± 0.08 & 0.65 ± 0.06 &  & 0.25 ± 0.12 & 0.93 ± 0.15 \\
ViT w/RoPE & Raw dMRI & MLP & Full & 47.40 ± 5.30 \% & 0.29 ± 0.07 & 0.46 ± 0.13 &  & -0.05 ± 0.18 & 1.08 ± 0.11 \\ \hline
\rowcolor[HTML]{FAFADC} 
Ours & Frozen latent & Linear & Head & 55.10 ± 3.50 \% & 0.44 ± 0.07 & 0.62 ± 0.11 &  & 0.18 ± 0.25 & 2.81 ± 1.12 \\
\rowcolor[HTML]{FAFADC} 
Ours & Frozen latent & MLP & Head & \textbf{64.70 ± 5.70 \%} & {\ul 0.50 ± 0.07} & 0.67 ± 0.02 &  & \textbf{0.38 ± 0.14} & \textbf{0.77 ± 0.07} \\
\rowcolor[HTML]{FAFADC} 
Ours & Last layer finetuned & MLP & Partial & 60.90 ± 4.10 \% & 0.49 ± 0.09 & 0.66 ± 0.07 &  & 0.32 ± 0.19 & 0.88 ± 0.15
\end{tabular}
}
\vspace{-15pt}
\end{table*}

Table~\ref{tab:hcpagdev_metrics} summarizes the results for the age prediction and sex classification tasks. The methods are grouped into handcrafted (HC) features (FA and MD, and RISH by segmented tract), fully trained methods (3D ResNet and ViTs trained from scratch), and our partially trained methods (frozen features obtained from our pretrained encoder combined with either a linear layer or an MLP, and our pretrained encoder with an MLP head with only the last encoding layer fine-tuned). The mean and standard deviation across the five cross-validation folds are reported. We observe a good overall performance of most methods with a high correlation coefficient ($\rho>0.89$), with the 3D ResNet and our finetuned pretrained model performing the best in terms of $\rho$ and MSE. In the sex classification task, possibly due to the lower number of samples, a simple method such as using \textbf{frozen latents} from our pretrained encoder with a linear layer performs better than finetuning the model. Overall, our method provides the second-best performance, behind a fully supervised method, while we used frozen, not-fully-supervised features. The good performance of even just linear probing on our frozen latents suggests that the learned representations already encode some structural information that is useful for both of these tasks (Some evidence of this structure is shown in Figure 8 in the supplement). 

Table~\ref{tab:adni_metrics} shows the results for the clinical tasks of cognitive score prediction and classification of cognitively normal (CN) versus mild cognitive impairment (MCI). For these tasks, the advantage of pretrained latent features becomes more apparent. Our frozen latents with an MLP head yield the best balanced accuracy (BA=64.7\%) in the CN vs MCI classification and also the best overall performance in terms of correlation coefficient and MSE in the ADAS prediction task. In these two tasks, our frozen latents and the HC features outperform other methods that are more data-intensive. The enhanced performance of our method with limited training data suggests that our pretrained encoder may have already internalized some microstructural patterns or features that are predictive of cognitive decline or disease transition. We also observe that our best performing method improves over ViT with standard RoPE applied on raw dMRI data. All metric comparisons between ViT with raw dMRI and our best performing method (except for MCI-AUROC and age-MSE) are statistically significant (paired t-test and Bonferroni correction).



\section{Discussion and Conclusions}

We have presented a novel transformer architecture designed to obtain general-purpose representations of diffusion MRI (dMRI). The model integrates mixed attention in both spatial and diffusion domains, along with a modified positional encoding specifically tailored to the geometry of the diffusion space (D-RoPE). Our findings indicate that the best performance is achieved by combining D-RoPE with alternating masking in the spatial and diffusion domains. As demonstrated in our reconstruction evaluation and ablation analysis, these components enable the model to encode relative information regarding different diffusion directions and b-values. Furthermore, the strong performance in FA and MD metrics derived from the reconstructed images suggests that sufficient microstructural information from the original images is encoded into the latent space, facilitating the reconstruction of those features. We also note that, throughout our tests, random directions were employed, and varying numbers of directions were utilized between pre-trained and downstream tasks, underscoring the model's ability to handle arbitrary diffusion acquisitions without the need to resample to a fixed acquisition scheme. However, a limitation of D-RoPE is that the distance we selected is not easily factorizable into keys and queries matrices, unlike standard RoPE, resulting in higher computational cost (For a sequence length $l$ and head dimension $h$, the cost of D-RoPE is $\mathcal{O}(lh^2)$ instead of $\mathcal{O}(lh)$ for regular RoPE). 

The learned representations we obtained were rigorously evaluated across various downstream tasks using additional datasets. Both the pre-computed features with a lightweight architecture on top and a partially fine-tuned pre-trained model performed competitively or even better than the various baselines, particularly in tasks with limited training data. In particular, our method improves the performance over a standard ViT with regular RoPE with Raw dMRI as input. This suggests our joint spatial and diffusion modeling is needed in the case of varying diffusion protocols to more fully extract relevant microstructural information.

Several studies have demonstrated a statistical relationship between various dMRI microstructural metrics and age~\cite{kochunov2012fractional,kumar2013brain,matijevic2021tract,conte2024white}. Given this evidence, the strong performance of the different tested methods is not unexpected. The sex‐classification task may be more challenging: dMRI microstructural sex differences have been reported during developmental ages~\cite{benavidez2024sex,lawrence2023white}, but large‐scale lifespan work~\cite{conte2024white} suggests that the nature and magnitude of these differences vary with age, which may complicate classification across different age groups. 

Conversely, the downstream tasks we evaluated in the ADNI dataset (classification of MCI and prediction of the ADAS) are considerably more challenging. 
Implementing actual classification models to detect MCI in larger cohorts has proven to be more difficult~\cite{wen2021reproducible} compared to classifying Alzheimer's Disease (AD). Our pretrained latent representations demonstrate an improvement in balanced accuracy when compared to traditional handcrafted features (64.70\% vs 61.90\%) similar to those used in the literature~\cite{maggipinto2017dti,wen2021reproducible}, indicating that the latent information likely encodes some microstructural patterns without the need to fit a biophysical model. Previous studies have shown correlations between dMRI metrics and various cognitive assessments~\cite{nir2013effectiveness,zavaliangos2019diffusion}. However, to the best of our knowledge, this is the first study to predict one of these cognitive assessments directly from dMRI volumes. The relatively high correlation coefficient we obtained ($\rho$=0.38) may indicate that the model can extract information from dMRI that is indicative of cognitive trajectories in this particular cohort.

The representation learning framework for dMRI presented here is significant for several reasons. Our model is capable of harmonizing analyses across cohorts with varying acquisition protocols, serving as a stepping stone toward the development of a foundation model for dMRI, which can be achieved by increasing the diversity and size of pretraining data. Furthermore, the latent information or attention maps may reveal new microstructural biomarkers that link white matter integrity to cognitive domains or other diseases. Additionally, the approach presented here could potentially be adapted for other applications, such as super-resolution, by integrating additional components like latent diffusion models. Finally, the representations may also play a crucial role in a future multi-modal model that integrates different MRI modalities or other medical information.

\section*{\textbf{Acknowledgments}}
This work was supported in part by National Institutes of Health grants AG089169, AG084471, DA057567 and AA021697, and the Stanford Institute for Human-Centered AI (HAI) Hoffman-Yee Award. A.C. receives research support from NIH grants R01 HL167974, R01HL169345, R01 AR077604, R01 EB002524, R01 AR079431, P41 EB 027060, P50 HD118632; Advanced Research Projects Agency for Health (ARPA-H) Biomedical Data Fabric (BDF) and Chatbot Accuracy and Reliability Evaluation (CARE) programs (contracts AY2AX000045 and 1AYSAX0000024-01); and the Medical Imaging and Data Resource Center (MIDRC), which is funded by the National Institute of Biomedical Imaging and Bioengineering (NIBIB) under contract 75N92020C00021 and through ARPA-H.
{
    \small
    \bibliographystyle{ieeenat_fullname}
    \bibliography{main}
}


\clearpage
\setcounter{page}{1}


\onecolumn
   {
        \centering
        \Large
        \textbf{\thetitle}\\
        \vspace{0.5em}Supplementary Material \\
        \vspace{1.0em}
       }

\subsection{Training and fine-tuning parameters}

The parameters used for the pretraining stage of the MAE are detailed in Table~\ref{tab:pret_param}, while those used for finetuning the last layer and the head of the MAE for the different downstream tasks are shown in Table~\ref{tab:ft_param}.

\begin{table}[h]
\caption{Parameters using for the pre-training stage of the MAE.}
\label{tab:pret_param}
\centering
\begin{tabular}{l|l}
\hline
Hyperparameter & Value \\ \hline
Number of transformer encoding Layers & 10 \\
Number of transformer decoding Layers & 3 \\
Number of convolutional decoding Layers & 3 \\
Latent dimension & 384 \\
Number of heads & 3 \\
Convolutional kernel size & 3 \\
optimizer & AdamW \\
optimizer momentum & $\beta_1$=0.9, $\beta_2$=0.999 \\
learning rate schedule & cosine \\
warmup epochs & 40 \\
start learning rate & 5.00E-05 \\
final learning rate & 1.00E-06 \\
weight decay schedule & cosine \\
initial weight decay & 0.04 \\
final weight decay & 0.4 \\
batch size & 4 \\
epochs & 300 \\ \hline
\end{tabular}
\end{table}

\begin{table*}[]
\caption{Parameters using for the partial finetuning of the MAE in different downstream tasks.}
\label{tab:ft_param}
\begin{tabular}{lllll}
\cline{2-5}
 & \multicolumn{4}{c}{Task} \\ \hline
hyperparameter & \multicolumn{1}{|l|}{Age prediction} & \multicolumn{1}{l|}{Sex Classification} & \multicolumn{1}{l|}{MCI classification} & ADAS prediction \\ \hline
optimizer & \multicolumn{1}{|l|}{AdamW} & \multicolumn{1}{l|}{AdamW} & \multicolumn{1}{l|}{AdamW} & AdamW \\
optimizer momentum & \multicolumn{1}{|l|}{$\beta_1=0.9$, $\beta_2=0.999$} & \multicolumn{1}{l|}{$\beta_1=0.9$, $\beta_2=0.999$} & \multicolumn{1}{l|}{$\beta_1=0.9$, $\beta_2=0.999$} & $\beta_1=0.9$, $\beta_2=0.999$ \\
learning rate schedule & \multicolumn{1}{|l|}{Constant} & \multicolumn{1}{l|}{cosine} & \multicolumn{1}{l|}{Constant} & cosine \\
start learning rate & \multicolumn{1}{|l|}{5.00E-05} & \multicolumn{1}{l|}{1.00E-05} & \multicolumn{1}{l|}{5.00E-05} & 5.00E-05 \\
final learning rate & \multicolumn{1}{|l|}{5.00E-05} & \multicolumn{1}{l|}{1.00E-06} & \multicolumn{1}{l|}{5.00E-05} & 1.00E-06 \\
weight decay schedule & \multicolumn{1}{|l|}{Constant} & \multicolumn{1}{l|}{Constant} & \multicolumn{1}{l|}{Constant} & cosine \\
initial weight decay & \multicolumn{1}{|l|}{0.01} & \multicolumn{1}{l|}{0.01} & \multicolumn{1}{l|}{0.05} & 0.05 \\
final weight decay & \multicolumn{1}{|l|}{0.01} & \multicolumn{1}{l|}{0.01} & \multicolumn{1}{l|}{0.05} & 0.01 \\
batch size & \multicolumn{1}{|l|}{6} & \multicolumn{1}{l|}{6} & \multicolumn{1}{l|}{6} & 6 \\
epochs & \multicolumn{1}{|l|}{50} & \multicolumn{1}{l|}{50} & \multicolumn{1}{l|}{50} & 20 \\
LoRA used? & \multicolumn{1}{|l|}{No} & \multicolumn{1}{l|}{No} & \multicolumn{1}{l|}{Yes} & Yes \\
rank (LoRA) & \multicolumn{1}{|l|}{-} & \multicolumn{1}{l|}{-} & \multicolumn{1}{l|}{12} & 12 \\
alpha (LoRA) & \multicolumn{1}{|l|}{-} & \multicolumn{1}{l|}{-} & \multicolumn{1}{l|}{12} & 12 \\ \hline
\end{tabular}
\end{table*}

\subsection{Additional evaluations}
\label{sec:add_eval}

\begin{figure*}[h]
    \centering
    \includegraphics[width=\linewidth]{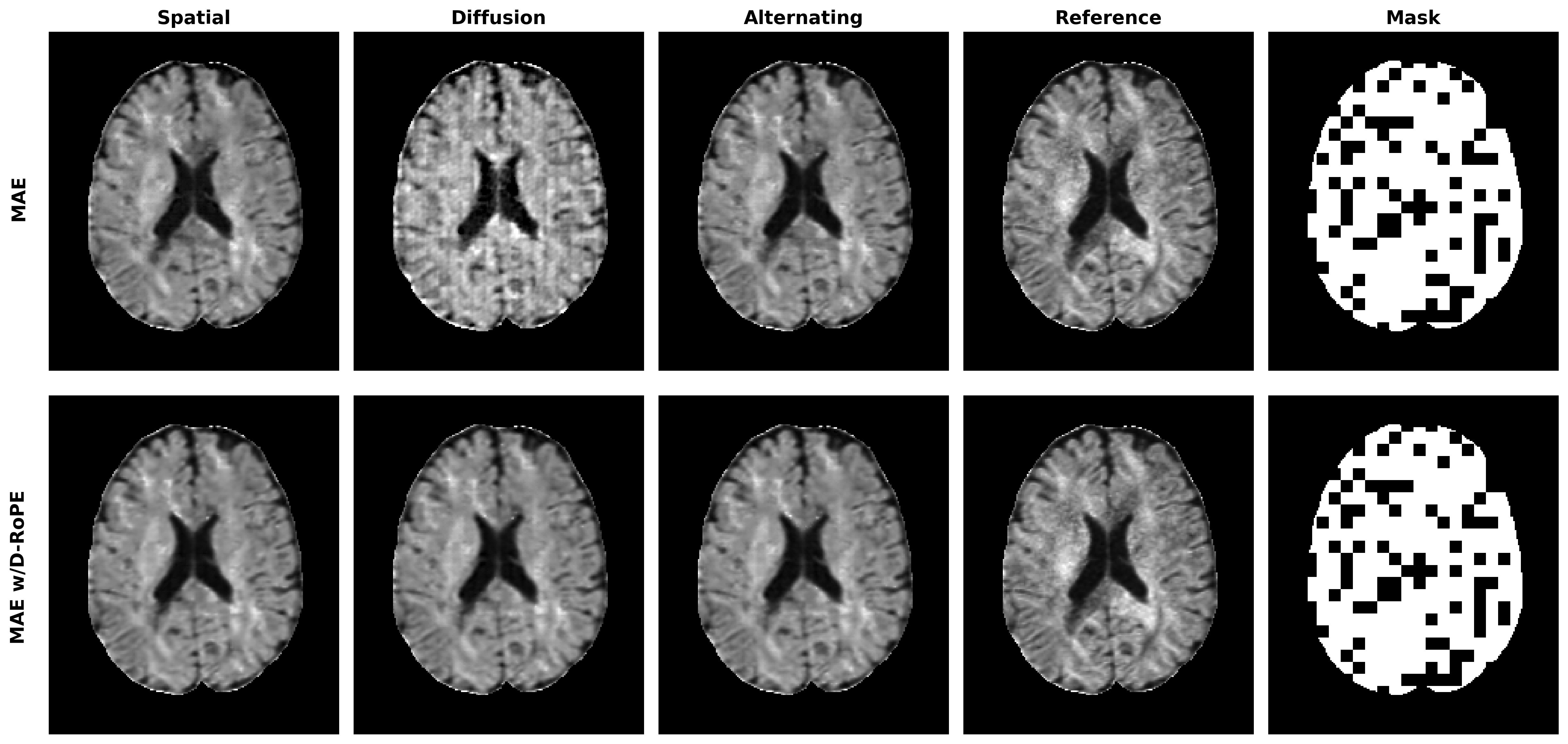}
    \caption{Qualitative evaluation of the reconstructions for a test subject for a b-value of 1000 $s^2/mm$. The reconstruction obtained from different masking strategies (Spatial, Diffusion and Alternating) and with and without D-RoPE are shown in the three left-most columns. The reference ground-truth image and the spatial mask that was used are shown in the two right-most columns.}
    \label{fig:b1000_recon}
\end{figure*}

\begin{figure*}[h]
    \centering
    \includegraphics[width=\linewidth]{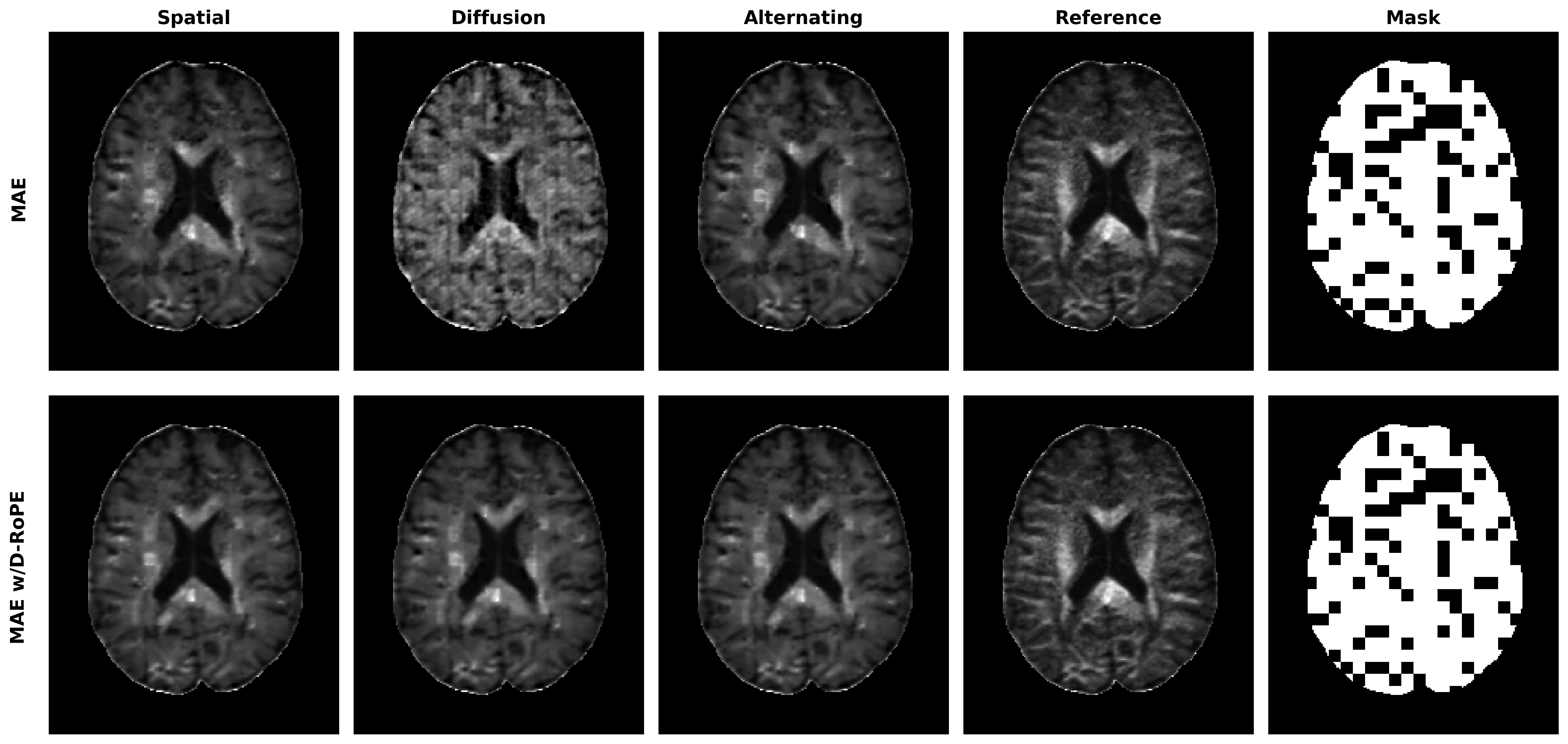}
    \caption{Qualitative evaluation of the reconstructions for a test subject for a b-value of 3000 $s^2/mm$. The reconstruction obtained from different masking strategies (Spatial, Diffusion and Alternating) and with and without D-RoPE are shown in the three left-most columns. The reference ground-truth image and the spatial mask that was used are shown in the two right-most columns.}
    \label{fig:b3000_recon}
\end{figure*}

Exemplary reconstructions from the different masking strategies (Spatial, Diffusion and Alternating) are shown in Figures~\ref{fig:b1000_recon} and ~\ref{fig:b3000_recon} for volumes with a b-values of 1000 $s^2/mm$ and 3000 $s^2/mm$, respectively. At 3000 $s^2/mm$, it is clearly observed how the version without D-RoPE and diffusion space masking captures well the structure but not the general intensity.

\begin{figure*}
    \centering
    \includegraphics[width=1\textwidth]
    {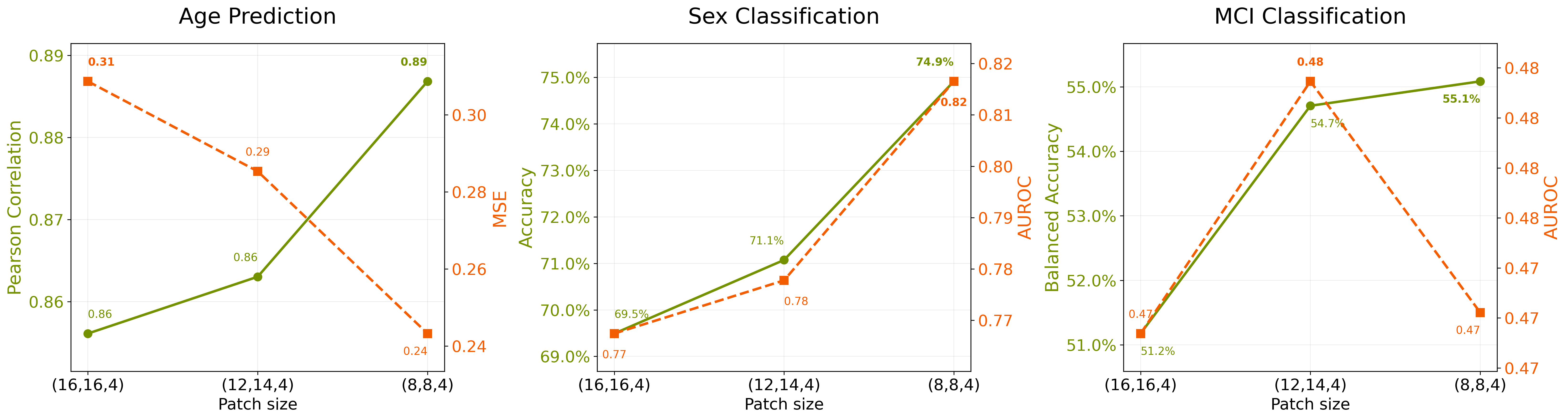}
    \includegraphics[width=1\linewidth]{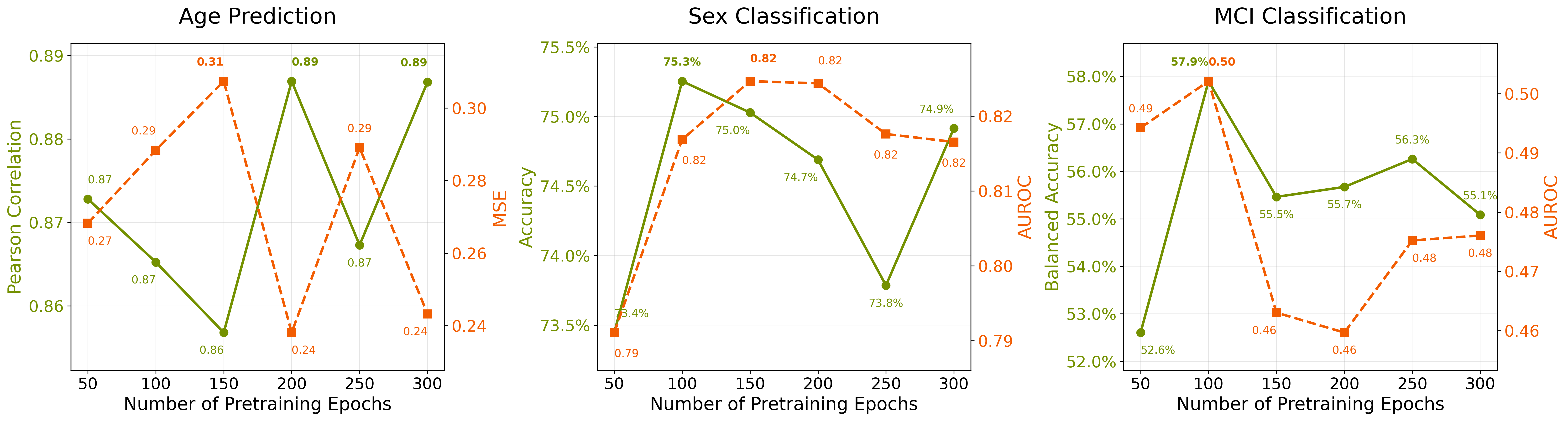}
    \caption{Linear probing performance of the frozen latents with different patch sizes (Top) and at different stages of pretraining (Bottom). A clear performance improvement in all downstream tasks is observed as one goes to smaller patch sizes. The optimal number of pretraining epochs seems to vary depending on the downstream task.}
    \label{fig:perepochs}
\end{figure*}

Figure~\ref{fig:perepochs} shows additional ablations with different patch sizes (Top) and at different stages of pretraining (Bottom). We evaluated patch sizes of (16,16,4), (12,14,4) and (8,8,4) and checkpoints of the model at 50, 100, 150, 200, 250 and 300 epochs in the pretraining stage. All other parameters are kept as in the baseline model (Table~\ref{tab:pret_param}). The evaluation was obtained with linear probing on top of the frozen latents as described in Section 4.3 and the plotted points correspond to the mean across the five-fold cross-validation setup. We observe a clear performance improvement in all downstream tasks as we utilize smaller patch sizes (e.g an improvement of approximately 5\% in sex classification going from the largest patch size to the smallest patch size). The results for different pretraining epochs vary more in their optimal points from one downstream task to another, with at least improvement in all metrics observed when going from 50 to 100 epochs.

To further show the effectiveness of our method in handling different protocols, we  designed an experiment where two different protocols of 12 random b-values and b-vectors are used, and latent representations are obtained with our pretrained model. A linear classifier was trained on Protocol 1 and then tested on either Protocol 1 or Protocol 2. The results are shown in Table~\ref{tab:comp_protocols}. We observe minimal variation between the results obtained from the representations of both protocols, suggesting that the latent representations are consistent across protocols and that the model is able to handle arbitrary protocols. 

\begin{table}[h]
\vspace{-5pt}
\caption{Comparison of linear probing on top of latent representations trained on protocol 1 and tested on this and a second different protocol. Minimal changes in performance are observed. \vspace{-10pt}}
\label{tab:comp_protocols}
\renewcommand{\arraystretch}{1.2} 
\centering
\begin{tabular}{ccccccc}
 &  & \multicolumn{2}{c}{Age prediction} &  & \multicolumn{2}{c}{Sex classification} \\ \cline{3-7} 
Trained on & Tested on & $\rho$ ($\uparrow$) & MSE ($\downarrow$) & \textbf{} & ACC ($\uparrow$) & AUROC ($\uparrow$) \\ \hline
Protocol 1 & Protocol 1 & 0.895 ± 0.010 & 0.22 ± 0.02 &  & 75.0 ± 1.2 \% & 0.827 ± 0.015 \\
Protocol 1 & Protocol 2 & 0.897 ± 0.007 & 0.21 ± 0.02 &  & 75.4 ± 2.0 \% & 0.828 ± 0.017
\end{tabular}
\end{table}

\subsection{Interpretability}

\begin{figure*}[h!]
    \centering
    \includegraphics[width=1\textwidth]{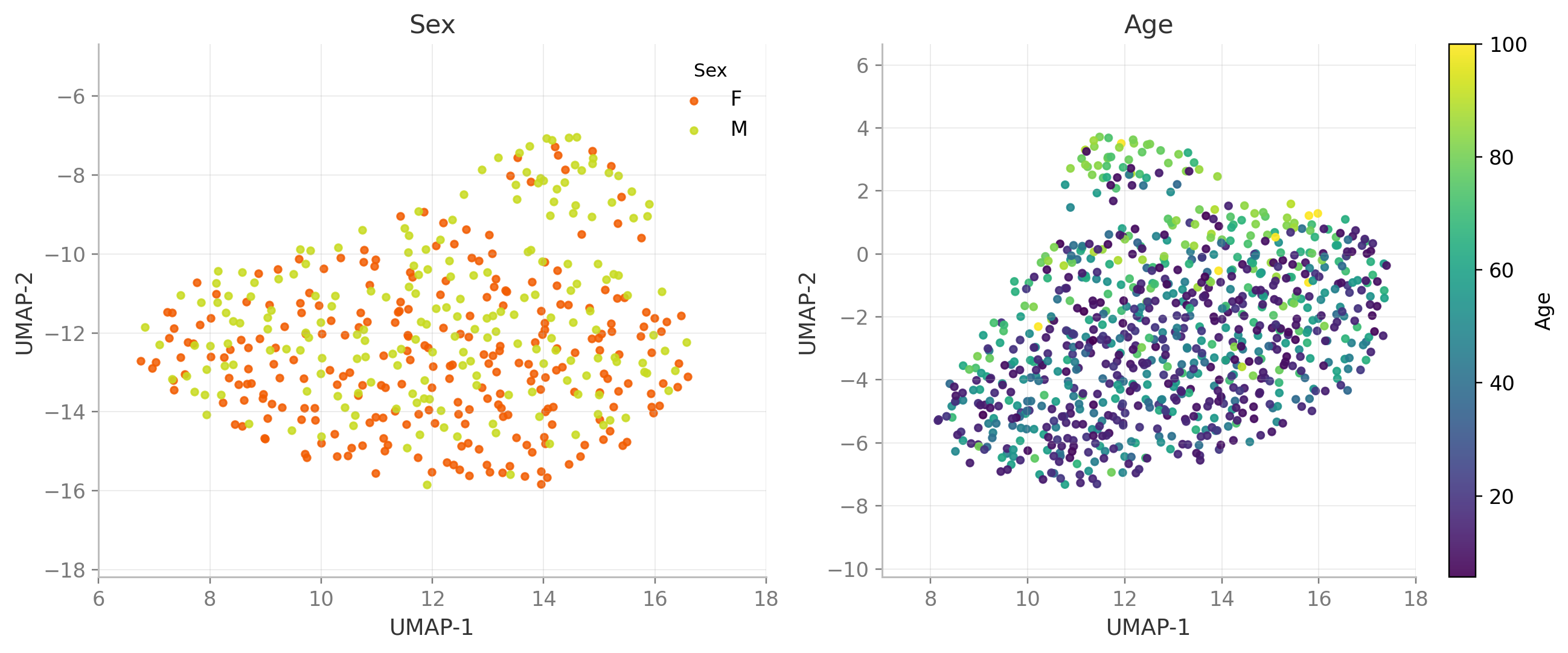}
    \caption{Two dimensional UMAP projections for the HCP-A samples colored by sex on the left and combined HCP-A and HCP-D samples colore by age on the right. We observed a structure of the latent space in which female subjects appear to be concentrated towards the lower part of the manifold, and in which there is a gradient from younger to older subjects going upwards in the manifold.}
    \label{fig:latents}
\end{figure*}

To explore the representations obtained with the pretrained model,  Figure~\ref{fig:latents} shows UMAP~\cite{mcinnes2018umap} projections for the HCP-A samples colored by sex on the left and combined HCP-A and HCP-D colored by age on the right. We observe certain qualitative structure of the latent space even without any finetuning. Female subjects appear to be concentrated towards the lower part of the manifold, and there is a gradient from younger to older subjects going upwards in the manifold.

\end{document}